\pdfoutput=1

\documentclass[11pt]{article}
\usepackage[T1]{fontenc}
\usepackage[hyphens]{url}
\usepackage[utf8]{inputenc}

\usepackage{EACL2023}

\usepackage{times}
\usepackage{latexsym}

\usepackage{microtype}

\usepackage{inconsolata}

\usepackage{times}
\usepackage{latexsym}
\usepackage{tikz,pgfplots}

\usepackage[T1]{fontenc}

\usepackage[utf8]{inputenc}
\usepackage{subcaption}
\usepackage{graphicx}
 
\usepackage{microtype}

\usepackage{inconsolata}

\usepackage{booktabs, makecell, multirow,comment}
\usepackage{siunitx}

\usepackage{arabtex}
\usepackage{utf8}
\setcode{utf8}
\usepackage{multirow,bigdelim}
\usepackage{blindtext}
\usepackage{svg}
\usepackage{acro}
\usepackage{pdfpages,caption}
\usepackage{todonotes,booktabs}

\DeclareAcronym{mt}{
short=MT,
long=machine translation
}

\DeclareAcronym{smt}{
short=SMT,
long= statistical machine translation
}
\DeclareAcronym{nlp}{
short=NLP,
long=natural language processing
}

\DeclareAcronym{atb}{
short=ATB,
long=Arabic Treebank
}
\DeclareAcronym{bpe}{
short=BPE,
long=Byte-Pair Encoding
}
\DeclareAcronym{oov}{
short=OOV,
long=Out-of-Vocabulary
}
\DeclareAcronym{lmvr}{
short=LMVR,
long=Linguistically Motivated Vocabulary Reduction
}

\DeclareAcronym{mcs}{
short=MCS,
long=morphological code-switching
}
\DeclareAcronym{cs}{
short=CS,
long=Code-switching
}
\DeclareAcronym{dea}{
short=EGY,
long= Egyptian Arabic
}
\DeclareAcronym{nmt}{
short=NMT,
long=neural machine translation
}
\DeclareAcronym{arzenss}{
short=ArzEnSEG,
long=ArzEn Surface Segmentation
}
\usepackage{xspace}
\newcommand{\BPE}{{BPE$_{joint}$}}
\newcommand{\RAW}{{raw}}
\newcommand{\MORFSRC}{{MORF$_{src}$}}
\newcommand{\MORFTGT}{{MORF$_{tgt}$}}
\newcommand{\FCSRC}{{FC$_{src}$}}
\newcommand{\FCTGT}{{FC$_{tgt}$}}
\newcommand{\LMVRSRC}{{LMVR$_{src}$}}
\newcommand{\LMVRTGT}{{LMVR$_{tgt}$}}
\newcommand{\LMVREGY}{{LMVR$_{src/egy}$}}
\newcommand{\LMVRJOINT}{{LMVR$_{joint}$}}
\newcommand{\MORPHASRC}{{MorphA$_{src}$}}
\newcommand{\MORPHATGT}{{MorphA$_{tgt}$}}
\newcommand\ATB{{MDMR$_{ATB}$}}

\newcommand\DT{{MDMR$_{D3}$}}

\newcommand{\LMVRMT}{{MT$_{LMVR}$}}
\newcommand{\BPEMT}{{MT$_{BPE}$}}
\newcommand{\ATBMT}{{MT$_{ATB}$}}
\newcommand{\MIXMT}{{MT$_{ATB+BPE}$}}

\newcommand{\hide}[1]{}

\title{Exploring Segmentation Approaches for Neural Machine Translation \\ of Code-Switched Egyptian Arabic-English Text}

\author{Marwa Gaser,$^1$ Manuel Mager,$^{2}$\thanks{~~Work done while at  the University of Stuttgart.}~ Injy Hamed,$^{3,4}$ \\
  {\bf Nizar Habash,$^4$ Slim Abdennadher,$^1$ Ngoc Thang Vu$^3$}\\
    $^1$The German University in Cairo, 
  $^2$AWS AI Labs \\
  $^3$University of Stuttgart, 
  $^4$New York University Abu Dhabi
  \\\texttt{\{marwa.saleh,slim.abdennadher\}@guc.edu.eg},
  \texttt{magerlm@amazon.com},
  \\\texttt{\{injy.hamed,nizar.habash\}@nyu.edu},
  \texttt{thang.vu@ims.uni-stuttgart.de}
  }

\begin{document}
\maketitle
\begin{abstract}
Data sparsity is one of the main challenges posed by code-switching (CS), which is further exacerbated in the case of morphologically rich languages. For the task of machine translation (MT), morphological segmentation has proven successful in alleviating data sparsity in monolingual contexts; however, it has not been investigated for CS settings. In this paper, we study the effectiveness of different segmentation approaches on MT performance, covering morphology-based and frequency-based segmentation techniques. We experiment on MT from code-switched Arabic-English to English. 
We provide detailed analysis, examining a variety of conditions, such as data size and sentences with different degrees of CS. Empirical results show that morphology-aware segmenters perform the best in segmentation tasks but under-perform in MT. Nevertheless, we find that the choice of the segmentation setup to use for MT is highly dependent on the data size. For extreme low-resource scenarios, a combination of frequency and morphology-based segmentations is shown to perform the best. For more resourced settings, such a combination does not bring significant improvements over the use of frequency-based segmentation.
 \end{list}
\end{abstract}



\section{Introduction}

\begin{figure}[t]
\centering
 \includegraphics[width=1\columnwidth]{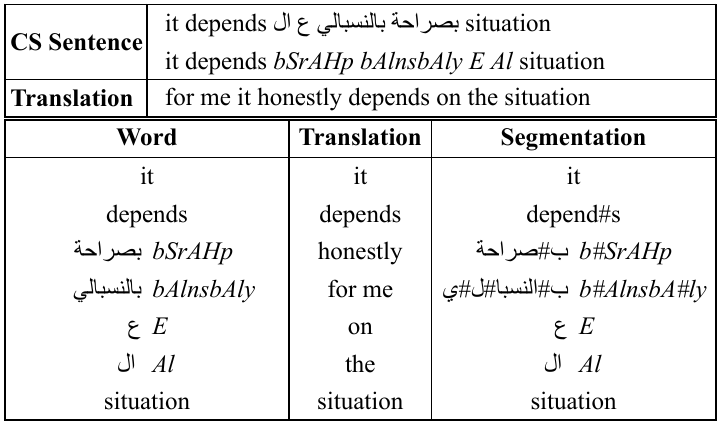}
   \caption{An example sentence with code-switching (CS) between English and Egyptian Arabic. The words are contrasted with their segmentations and English translations. Arabic words are paired with their transliterations in the Buckwalter scheme \cite{Habash:2007:arabic-transliteration}.} 
  \label{fig:ArzEnSEGs_ex}
\end{figure}

\ac{cs}, i.e. the alternation of language in text or speech, has been gaining world-wide popularity, due to several reasons, including globalization and immigration. While this has been met with a growing interest in the NLP field to build systems that can handle such mixed input, work on CS \ac{mt} 
is still considered in its infancy, where only a few language pairs have been investigated \cite{SK05,DKS18,MLJ+19,XY21,hamed2022investigating}.

%

In this work, we focus on the CS \ac{dea}-English (EN) 
language pair, as we observe its usage is becoming more common. Besides being prevalent amongst Egyptian migrant communities, it is also commonly used in Egypt due to the increase in international schooling systems and educational advancements. 
We identify three main challenges for CS \ac{mt}.
First is {\bf data sparsity}, a challenge common to many CS language pairs because of limited parallel corpora containing commissioned translations of CS text \cite{ccetinouglu2016challenges,SS20,TKJ21,hamed2022arzenST,CAS+22}.
%
Second is Egyptian Arabic {\bf morphological richness}, which further exacerbates the data sparsity situation \citep{habash2012conventional,habash2012morphological}. 
%
%
Third, since the matrix language (EGY) is morphologically rich, CS occurs at three {\bf CS  levels}:  on the boundaries of sentences (inter-sentential CS), between words (intra-sentential CS), and within words, i.e., \ac{mcs}. This mix of types of CS raises the question of how to handle them all in the same system. These challenges are further illustrated in Figure~\ref{fig:ArzEnSEGs_ex}.
\newpage 


A common solution to handle data sparsity for \ac{mt} of morphologically rich languages is morphological segmentation \cite{maiodeh,lmvr,gronroos2020morfessor}. However, this has not been investigated for CS. 
In this paper, we explore a wide range of segmentation approaches, covering unsupervised morphology-based segmenters, unsupervised frequency-based segmenters, and supervised morphology-based segmenters. This work aims to answer the following research questions (RQs): 
\begin{itemize}
\item \textbf{RQ1:} Which segmentation setup 
performs the best in the downstream \ac{mt} task across different training sizes?
\item \textbf{RQ2:} Does the effectiveness of the different segmenters in the MT task differ according to the \ac{cs} type of the source sentence? 
\item \textbf{RQ3:} Is there a correlation between a more morphologically correct segmentation and \ac{mt} performance? 
\end{itemize}

While our results show that there is no correlation between correct morphological segmentation and MT performance, we find that the performance ranking between the MT systems varies across different training data sizes and sentence types (monolingual vs. code-switched).
\hide{
we find that there is an interplay between the different segmenters, where the ranking varies across different training data sizes and sentence types (monolingual vs. code-switched).
}
We show that applying a combination of supervised morphology-based and unsupervised frequency-based segmentations consistently gives best results, with statistical significance under low data sizes. While common-wisdom suggests that \ac{bpe} is the best approach, our experiments highlight the importance of integrating morphological knowledge in the case of extreme low-resource settings. We believe that the insights and methodology we follow will be useful to researchers working with low-resource languages. An additional contribution of our research is the creation of a gold standard morphologically annotated \ac{cs} Egyptian Arabic-English dataset which we 
make publicly available.\footnote{\url{http://arzen.camel-lab.com/}}




The paper is organized as follows. Section~\ref{sec:previous_work} discusses related work. Section~\ref{sec:data_annotation} describes the 
dataset we annotated. Section~\ref{sec:segmentation} describes and evaluates
the different segmenters used. Section \ref{sec:MT} describes and evaluates the various \ac{mt} systems. In Section~\ref{sec:discussion}, we answer our research questions. 


\section{Related Work} \label{sec:previous_work}
Several researchers have investigated the effect of applying different morphological and agnostic segmentation approaches on the \ac{mt} performance for monolingual languages. \citet{roest2020machine,saleva21} show that unsupervised morphology-based segmentation like \ac{lmvr} \citep{lmvr}, Morfessor \citep{morfessor}, and FlatCat \citep{flatcat} for Nepali--English, Sinhala--English, Kazakh--English, and Inuktitut--English language pairs show either no improvement or no significant improvement over the agnostic \ac{bpe} segmentation \citep{sennrich16} in translation tasks. Meanwhile, \citet{mager2022bpe} and \citet{lmvr} show that for 
polysynthetic and highly agglutinative languages, unsupervised morphology-based segmentation outperforms BPEs \citep{sennrich16} in \ac{mt} tasks in both directions. Nevertheless, applying  \ac{bpe}s on top of morphology-based segmentation for Turkish--English, Uyghur--Chinese, and  Arabic--English has shown to bring improvements over solely using BPEs or morphology-based segmentation for neural \ac{mt} task \citep{pan2020morphological,tawfik2019morphology}. 
A similar result was achieved by \citep{ortega2020neural}, using a morphological guided BPE for polysynthetic languages.
However, \citet{maiodeh} show that such an approach is beneficial in the case of \ac{smt}, and does not improve results for \ac{nmt}. For other \ac{nlp} tasks, \citet{al2015effect} show that utilizing word segmented Arabic dataset leads to improvements in text classification task over utilizing unsegmented dataset in terms of accuracy, precision, recall, and F-measure.

%
As for work on CS MT, there are many efforts
\cite{SK05,DKS18,MMD+19,MLJ+19,SZY+19,TKJ21,XY21,CAS+22,hamed2022investigating}. 
To the best of our knowledge, none of these efforts presented an extensive comparison covering different segmentation techniques. With regards to the languages covered, only \citet{MLJ+19} worked on CS Arabic-English. However, since they used carefully edited UN documents, the text only included the Modern Standard Arabic variety, and contained limited types of CS.

With regards to similar corpora, \citet{balabel2020cairo} annotated CS Egyptian Arabic-English data 
\cite{hamed2018collection} 
with tokenization (canonical segmentation), lemmatization, and POS tags. 
However, their corpus does not contain translations.


\section{Data} \label{sec:data_annotation}
\subsection{Pre-existing Datasets}
We use the \textit{ArzEn} parallel corpus \cite{hamed2022arzenASR,hamed2022arzenST}, which
consists of speech transcriptions gathered through informal interviews with bilingual Egyptian Arabic-English speakers, as well as their English translations. The corpus consists of 6,213 sentences, where 4,154 (66.9\%) are code-mixed, 1,865 (30.0\%) are monolingual Arabic, and 194 (3.1\%) are monolingual English. 
Among the code-mixed sentences, there are 1,781 (28.7\%) sentences with morphological code-switching.
We follow the predefined dataset splits, containing 3,341 (53.8\%), 1,402 (22.6\%), and 1,470 (23.7\%) sentences for train, dev, and test sets, respectively. For training purposes, we also use 308k monolingual parallel sentences obtained from MADAR \citep{bouamor2018madar} and the following \textit{LDC} corpora: \citep{LDC97T19,LDC2002T38,LDC2002T39,ldc2017t07,LDC2021T15,LDC2012T09,song2019}. The preprocessing steps we apply are outlined in Appendix~\ref{sec:appendix-preprocessing}. 
We use \textit{ArzEn} train set as well as the monolingual parallel corpora to train both the segmenters and MT systems. For tuning and testing the MT systems, we use the \textit{ArzEn} dev  and test sets. For tuning and testing the segmenters, we annotated a new dataset, discussed next.

\hide{
We use \textit{ArzEn} parallel corpus provided by \citep{hamed2022investigating}. The corpus
We use \textit{ArzEn} train set as well as the monolingual parallel corpora to train both the segmenters and MT systems. For tuning and testing the MT systems, we use \textit{ArzEn} dev  and test sets. For tuning and testing the segmenters, we use a subset form Arzen's dev set to develop a code-switched morphologically annotated dataset, which we refer to as \textit{\ac{arzenss}}, described in the following section.
}
\subsection{A New Dataset: \ac{arzenss} Corpus}
\begin{table}[t]
\centering
\small
\setlength{\tabcolsep}{2pt}
\begin{tabular}{|l|l|l|l|}
\hline
{\bf Case} & {\bf Stem} & {\bf Ending}& {\bf Example}\\
\hline
Irregular & modified & Irregular: es & monki+es\\
Irregular & modified & Regular: s,ed,ing,en & car+ing\\
Irregular & modified & Irregular: <nil> & went \\
Irregular& unmodified & Irregular: es & church+es \\
Regular& unmodified &Regular: s,ed,ing,en & car+s \\
\hline
\end{tabular}
\caption{The ordered list of rules we follow to segment the English words.
}
\label{english_rules}
\end{table}

To facilitate our research, we created a code-switched Egyptian Arabic-English morphologically annotated dataset which we use for tuning and testing. The dataset comprises the first 500 lines of \textit{ArzEn} dev set. Unlike \citet{balabel2020cairo}, we opt for surface form segmentation to allow for evaluating the segmenters. We also opt for extending \textit{ArzEn} dataset as it contains translations and is used in our MT experiments.

For Arabic word segmentation, we use the \ac{atb} segmentation scheme \citep{maamouri2004penn,habash2010introduction}. We choose this scheme as it is the standard tokenization scheme used in different treebanks \cite{maamouri2004penn,Maamouri:2012:arz,taji2017universal,habash2022camel}. It has also shown to be efficient in \citet{maiodeh} and has demonstrated its competitiveness in \citet{habash2013morphological}. 

For English word segmentation, we follow five rules in sequential order depending on whether the word has a regular or irregular stem and whether the word has a regular or irregular ending.
Table~\ref{english_rules} exhibits the five English rules we follow in order. 

All annotation decisions were made in context by two bilingual speakers who collaborated on initial annotations and quality checks. Figure~\ref{fig:ArzEnSEGs_ex} presents an annotation example.
We divide the sentences randomly into dev and test sets (250 sentences each). In Table~\ref{table:ArzEnSS_stats}, we display statistics about \textit{\ac{arzenss}}.

\hide{
\begin{table}[t]
\centering
\small
\begin{tabular}{|l|r|r|}
\cline{2-3}
\multicolumn{1}{c|}{} & {\bf EGY}&{\bf EN}\\
\hline
Test  &3,414&501 \\
\hline
Dev &3,069& 567\\
\hline
Total&6,483&1,068 \\
\hline
\end{tabular}
\caption{
We display the total number of words in the dev and test sets of\textit{ \ac{arzenss}}.
}\label{table:splits}
\end{table}
}
\begin{table}[t]
\centering
\small
\begin{tabular}{|l|r|r|}
\cline{2-3}
\multicolumn{1}{c|}{} & \textbf{EGY} &  \textbf{EN}\\
\hline
Test  Words &3,414&501 \\
Dev Words &3,069& 567\\
Total Words  & 6,483&1,068\\
\hline\hline
Total Segmented Words&1,206 & 146\\
Total Morphs&7,911&1,214 \\
Total Unique Morphs& 1,192& 432\\
\% of Total Segmented Words  & 18.6\%& 13.7\% \\
Morphs/Word &1.220 &1.137\\
Maximum Morphs per Word & 5 &2\\
\hline 
\end{tabular}
\caption{
 Statistics on \textit{\ac{arzenss}} corpus.
}\label{table:ArzEnSS_stats}
\end{table}
 
\section{Segmentation Experiments}\label{sec:segmentation}
\subsection{Experimental Setup}
\begin{figure*}[ht]
\centering
 \frame{\includegraphics[width=2\columnwidth]{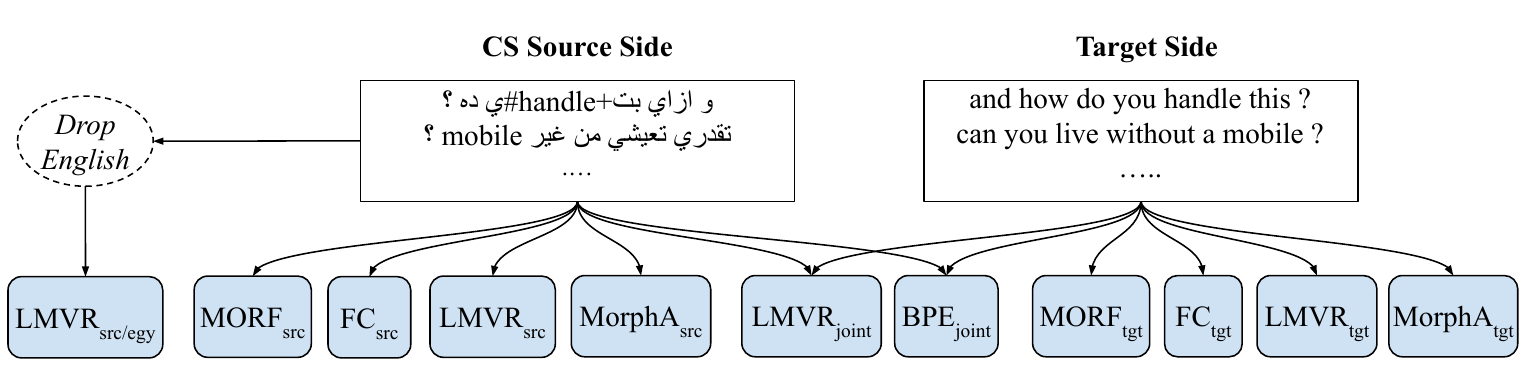}}
   \caption{The unsupervised segmentation models we study in this paper and their training data dependencies. We use four systems: Morfessor (MORF), FlatCat (FC), LMVR, and MorphAGram (MorphA). The subscripts specify the training data: source (src), target (tgt), source+target (joint), and source without English, i.e., Egyptian, (src/egy).}
  \label{fig:training_unsupervised_segmenters}
  \vspace{-0.3cm}
\end{figure*}

We explore three categories of segmenters: unsupervised morphology-based, unsupervised frequency-based, and supervised morphology-based segmentation. For the unsupervised morphology-based segmenters, we use MorphAGram in addition to three segmenters from the Morfessor family: Morfessor, LMVR, and FlatCat. For unsupervised frequency-based segmenters, we use BPE. Figure~\ref{fig:training_unsupervised_segmenters} summarizes the process of training these segmenters. For the supervised morphology-based segmenters, we use MADAMIRA \citep{madamira}, 
where we exploit the segmentation schemes designed for Egyptian Arabic.

\subsection{Segmentation Systems}
\label{subsec:segmentation_systems}

In this section, we introduce the segmentation systems used for the study. Details about the hyperparameter tuning for each system family can be found in Appendix~\ref{appx:segmenters}.
The different segmentation models and their training dataset are displayed in Figure~\ref{fig:training_unsupervised_segmenters}.

\paragraph{Morfessor Family}
We exploit three Morfessor family tools for unsupervised morphology-based segmentation in this research: Morfessor, \citep{morfessor}, FlatCat \citep{flatcat}, and \ac{lmvr} \citep{lmvr}. 

\textbf{Morfessor} is a morphological-based segmentation model which we train in an unsupervised manner. Three components form the system: the model, the cost function, and the training and decoding algorithms \citep{virpioja2013morfessor}. The model is mainly concerned with the grammar and lexicon where the latter holds the attributes of the subwords and the grammar controls how these subwords are combined to form the word. Morfessor's grammar assumes that the subwords that form the word are independent of each other and that a word has at least one subword.

\textbf{FlatCat} is a variant of Morfessor which we also train in an unsupervised manner. Even though FlatCat builds on Morfessor and shares the same model component, they differ in their morphotactics (the set of rules that determine how the word's morphemes are arranged).
FlatCat morphotactics is based on the Hidden Markov model \citep{baum1966statistical} which considers context. On the contrary, Morfessor's morphotactics algorithm is based on a unigram model which is not context-sensitive.

\textbf{\ac{lmvr} }is a morphology-based segmenter that is built upon FlatCat and we train in an unsupervised manner. Nonetheless, \ac{lmvr} takes into consideration the desired segmentation output vocabulary size during training. 

For each tool, two models are generated; one trained on the source side; thus capable of segmenting CS data, and the other trained on the 
target side of the training data; thus capable of segmenting English data only. We add a \textit{src} and \textit{tgt} subscript to the segmenters' names to distinguish between both settings.
Hence, \MORFSRC{}, \FCSRC{}, and \LMVRSRC{} resemble Morfessor, FlatCat, and \ac{lmvr} respectively, where the segmenters are trained on the source side. \MORFTGT{}, \FCTGT{}, and \LMVRTGT~resemble the segmenters trained on the target side. 

\paragraph{MorphAGram} We also include in this study the unsupervised morphology segmenter MorphAGram \citep{eskander2020morphagram} which is based on Adaptor Grammars.
We use the \textit{PrStSu+SM} grammar, which represents a word as a sequence of prefixes followed by a stem then a sequence of suffixes, in the unsupervised \textit{Standard} learning setting to train the segmenters.   

\paragraph{\ac{bpe}}
The SentencePiece \citep{kudo2018sentencepiece} implementation of \ac{bpe} \citep{gage1994new,sennrich16} is a frequency-based unsupervised segmenter. 
We train the \ac{bpe} model jointly, on the concatenation of the source and target sides of the training parallel corpus. Previous work has shown that this approach is better suited for low resource settings \cite{flores}. 
We refer to our joint \ac{bpe} segmenter 
as \BPE.

\paragraph{MADAMIRA}
For supervised morphology-based segmenters, we use MADAMIRA's Egyptian Arabic model \citep{madamira}, which was trained on the Egyptian Arabic Treebank (parts 1 through 6) \citep{Maamouri:2012:arz}. 
Specifically, we use MADAMIRA's \textit{\ac{atb}\_BWFORM} and \textit{D3\_BWFORM} schemes, 
henceforth \ATB~and \DT, respectively.
%
Both 
schemes apply Alif/Ya normalization and segment the Arabic clitics. {\DT} splits the Arabic definite article \<ال> \textit{Al} (the), while {\ATB} does not.

\subsection{Segmentation Results}\label{subsec:segmentation_results}
To evaluate the performance of the
segmenters, we use EMMA F1 score \citep{spiegler2010emma}. Results in Table~\ref{table:EMMA_scores}, reported on \textit{\ac{arzenss}} test set, show overall and language-specific scores.



\begin{table}[t]
\small
\centering
\begin{tabular}{|c|c|c|c|c|}
\cline{2-4}
\multicolumn{1}{c}{} &\multicolumn{3}{|c|}{{\bf EMMA F1 Score}}   \\\hline 
  
{\bf Segmenter} & {\bf EGY} &{\bf EN} & {\bf All} \\
   \hline\hline 
 raw &0.806& 0.953&0.838\\\hline 
 \hline
 \MORPHASRC &0.682& 0.942&  0.737 \\
 \MORFSRC & 0.814& 0.888  & 0.832  \\
  \FCSRC &  0.821&0.961& 0.851\\
   \LMVRSRC  &0.836& 0.961&0.863  \\
     \LMVREGY & 0.838 &  0.953& 0.863   \\\hline

\MORPHATGT  & 0.806 &0.953& 0.838  \\
 
\MORFTGT & 0.147 &0.951 & 0.327 \\
 \FCTGT  & 0.806&0.952 & 0.838\\

 \LMVRTGT  & 0.806&0.966 & 0.842  \\\hline

 \LMVRJOINT  &0.841& 0.963 &\textbf{0.868 }  \\

\hline\hline 
 \BPE & 0.678&  0.814&\textbf{0.707}  \\
 
 \hline\hline 

\ATB&0.935 &  0.953&\textbf{0.939}  \\
 \DT   &0.868& 0.953& 0.887  \\
 \hline
\end{tabular}

\caption{
EMMA F1 score calculated on \textit{\ac{arzenss}} test set for the raw data as well as the segmented data using the different segmenters. The Arabic gold segmentation is based on the \ac{atb} segmentation scheme. We show the overall score (All) and language-specific scores calculated on the Egyptian Arabic (EGY) and English (EN) words separately. 
Segmenter names with a \textit{src}, \textit{tgt}, and \textit{joint} subscripts represent segmenters that are trained on the source, target, and source+target sides respectively. The best performing segmenters from each category are highlighted in bold.}
\label{table:EMMA_scores}
\end{table}

\paragraph{Unsupervised morphology-based segmentation} Results show that \ac{lmvr} outperforms the other unsupervised morphology-based segmenters in terms of segmenting 
Arabic and English words.
We perform further experiments where we train 2 additional models:
i) a model trained jointly on the concatenation of the source and target sides of the parallel corpus,
and ii) a model trained on the Arabic words only in the source side (where English words are dropped). 
Therefore, the former model is capable of segmenting both languages, while the latter is only tailored for segmenting Arabic words. We perform these experiments using \ac{lmvr}, given that it outperforms the other segmenters. We refer to these models as \LMVRJOINT~and \LMVREGY~respectively, as outlined in Figure~\ref{fig:training_unsupervised_segmenters}. 
Results show that joint training provides best EMMA scores. 


\paragraph{Supervised morphology-based segmentation} As shown in Table~\ref{table:EMMA_scores}, both supervised morphology-based segmenters {\ATB} and {\DT} outperform all other segmenters. Their superiority in segmenting Arabic is expected, as they are trained on human-annotated data and hence are capable of generating infrequent morphemes. Additionally, MADAMIRA has a morphological analyzer embedded in it, which in turn enriches the inspection of Arabic words
prior to segmentation. 
Higher EMMA scores are reported for {\ATB} over {\DT}, which is also expected, as \textit{\ac{arzenss}} is segmented following the \ac{atb} scheme. 

\paragraph{Unsupervised frequency-based segmentation} As expected, \BPE~performs the worst in the morphology-based segmentation task, as it is designed for agnostic segmentation for the purpose of improving downstream tasks.

\paragraph{Further analysis}
We surprisingly find that \MORPHATGT~outperforms \MORPHASRC~on 
Arabic words and \FCSRC~outperforms \FCTGT~on English 
words. Therefore, we conduct 
an 
internal analysis where we look into the percentage of over and under segmentations.\footnote{Over segmentation is a term we use to indicate that the word gets segmented to more morphemes compared to the gold standard segmentation. Meanwhile, under segmentation is a term we use to convey that the word is segmented into fewer morphemes than the gold standard segmentation.} In 
Appendix~\ref{sec:appendix-SegmentersPerformanceAnalysis}, we present the number of under and over segmented words for each segmentation approach. 
Our analysis shows that \MORPHASRC~over segments 25\% of the Arabic words. We observe that in 20\% of these over segmentation cases, the Arabic definite article is segmented. For example, the word \<الكتب> \textit{Alktb} `the books' is segmented to \<كتب>\#\<ال> \textit{Al\#ktb} which is considered valid in segmentation schemes like D3. However, since we use the \ac{atb} scheme in \textit{\ac{arzenss}} annotation, the EMMA system penalizes the \MORPHASRC~segmenter and rewards the \MORPHATGT~segmenter which leaves most of the Arabic words and the definite article unsegmented.  Another case is the segmentation of affixes, which is not done in ATB. For example, 16\% of the over segmentation cases are separation of the Ta-Marbuta (feminine nominal ending) in Arabic words. The rest of the cases are grammatically incorrect  segmentations. 
\FCTGT~is also shown to under segment around 17\% more English words compared to \FCSRC~which can contribute to worse scores. 
We also observe that \MORFTGT~performs significantly worse than the other segmenters when segmenting Arabic words, despite the fact that 81\% of the Arabic words do not require segmentation. Internal analysis shows that \MORFTGT~over segments the Arabic words to the character level in an attempt to extract the underlying morphology of  Egyptian Arabic, which it was not trained on.

\section{Machine Translation Experiments}\label{sec:MT}
Since no previous research investigates the best segmentation technique for \ac{nmt} of 
the code-switched 
Egyptian Arabic--English language pair, we explore 
training \ac{nmt} models using the various segmentation setups discussed in Section~\ref{sec:segmentation} to answer RQ1. 
Moreover, we analyze the performance of the top-performing \ac{mt} systems on different types of \ac{cs} sentences to answer RQ2.
Afterward, we compare the MT scores against the EMMA F1 scores discussed in Section~\ref{subsec:segmentation_results} to answer RQ3.

\subsection{Experimental Setup}
We train Transformer models for our \ac{mt} systems using Fairseq \cite{fairseq} on a single GeForce RTX 3090 GPU. We use the hyperparameters from the FLORES\footnote{FLORES hyperparameters outperform \citet{vaswani2017attention} for our code-switched pair by +0.4 chrF2++ points.} benchmark for low-resource \ac{mt} \citet{flores}, which we list in Appendix~\ref{sec:appendix-MThyperparameters}. Afterwards, we evaluate the \ac{mt} models on \textit{ArzEn}'s dev and test sets using chrF2++ \citep{popovic-2017-chrf}.\footnote{We use sacreBLEU's \citep{sb} implementation of chrF2++.} We choose chrF2++ over BLEU \citep{papineni2002bleu} 
as it rewards partially correct translations which makes it a convenient choice for our research, and because chrF has shown to have higher correlation with human judgments over BLEU \citep{ship}.

\subsection{Machine Translation Systems}



\begin{table}[t]
\centering
\small
\setlength{\tabcolsep}{1pt}
\begin{tabular}{|c|c|c|c|l|}
\hline
\multicolumn{3}{|c|}{{\bf Segmentation}} & \multicolumn{2}{c|}{\bf chrF2++} \\\hline 
\multicolumn{2}{|c|}{{\bf Source}} & {\bf Target} & \multicolumn{2}{c|}{}\\  \cline{1-3}
 {\bf EGY} & {\bf EN} & {\bf EN} &  \multicolumn{1}{c}{{\bf dev}} &  \multicolumn{1}{c|}{{\bf test}}       \\ \hline\hline
\multicolumn{2}{|c|} \RAW & \RAW & 47.1 & 49.9  \\
\hline\hline
\multicolumn{5}{|c|}{\it Unsupervised Morphology-based Segmenters}\\\hline
\multicolumn{2}{|c|}\MORPHASRC &  \MORPHATGT & 47.0 & 49.7\\
\multicolumn{2}{|c|}\MORFSRC  & \MORFTGT  & 47.4 & 50.8\\
\multicolumn{2}{|c|}\FCSRC  &   \FCTGT  & 47.2 & 50.6\\
\multicolumn{2}{|c|}\LMVRSRC  &   \LMVRTGT & 48.3 & 51.7\\
\multicolumn{2}{|c|}\LMVRJOINT  &   \LMVRJOINT & 48.8&52.5\\\hline
\multicolumn{1}{|c|}\LMVREGY  &  \LMVRTGT &  \LMVRTGT & {\bf 48.9}& \textbf{52.9}\\
\multicolumn{1}{|c|}\LMVRSRC  &  \LMVRTGT &  \LMVRTGT & 48.8 & \textbf{52.9}\\
\multicolumn{1}{|c|}\LMVREGY  &  \LMVRSRC &  \LMVRTGT & 48.5 &52.0\\
\hline\hline
\multicolumn{5}{|c|}{\it Frequency-based Segmenters}\\\hline

\multicolumn{2}{|c|}\BPE &  \BPE & {\bf 50.1}&\textbf{53.7}\\
\multicolumn{2}{|c|}\BPE &  \RAW & 47.4 & 50.8\\
\multicolumn{2}{|c|}\RAW &  \BPE & 44.3 & 46.9\\
 \hline\hline
 \multicolumn{5}{|c|}{\it Supervised Morphology-based Segmenters}\\\hline

{\ATB}  &  {\RAW} &  {\RAW} & {\bf 48.8} & \textbf{52.1}\\
{\DT}  &  \RAW &  \RAW & 47.9 & 51.1\\
\hline\hline
\multicolumn{5}{|c|}{\it Combination Segmenters}\\\hline
{\ATB+\BPE}  &  \BPE &  \RAW & 46.5 & 50.1\\
{\ATB+\BPE}  &  \BPE &  \BPE & {\bf 50.2} & \textbf{53.8} \\
{\DT+\BPE}  &  \BPE &  \RAW &  46.9 & 50.7\\
{\DT+\BPE}  & \BPE &  \BPE &  49.8 & 53.3\\
\hline
\end{tabular}
\caption{The chrF2++ results of our \ac{nmt} systems with different segmentation combinations on \textit{ ArzEn}'s dev and test sets. Numbers highlighted in bold show the best performing system in each category.}
\label{table:chrf2_scores_dev}
\end{table} 

We experiment with different categories of segmentation setups. Table \ref{table:chrf2_scores_dev} shows all the different setups that we explore. 
See Table \ref{table:training_time_mt} in Appendix~\ref{sec:appendix-MThyperparameters} for training time.

For the \textbf{unsupervised morphology-based segmentations}, we use 
MorphAGram, Morfessor, FlatCat, and LMVR to segment the source/target sides of the parallel corpus, where the segmenters were trained on each side separately (see Figure~\ref{fig:training_unsupervised_segmenters}). 
For the best performing segmenter, we further investigate the best training setting, where we investigate using segmenters trained only on Arabic words on the source side as well as segmenters that are trained jointly on both sides.


For 
the \textbf{supervised morphology-based segmentations}, we only follow one approach and that is segmenting the source side using \ATB~or \DT~segmenters. This causes the English words to be left unsegmented. 

For the \textbf{unsupervised frequency-based segmentations}, we exploit the jointly trained model, \BPE, to segment the source side only, target side only, or both sides of the 
parallel corpus.


Finally, inspired by the work of \citet{maiodeh}, 
we explore \textbf{combinations} between \ac{bpe} and supervised morphology-based segmenters. 
As shown in Table~\ref{table:chrf2_scores_dev},  
for the source side, we apply 
\BPE~on top of segmentations provided by either \ATB{} or \DT{}. For the target side, 
we either leave it in the \RAW~format or apply \BPE~. 




\subsection{Machine Translation Results}

Table~\ref{table:chrf2_scores_dev} shows the different \ac{mt} systems and their performance on \textit{ArzEn}'s dev and test sets.


Amongst the \textbf{unsupervised morphology-based segmenters}, 
\ac{lmvr} outperforms the other segmenters. We find that training language-specific segmenters (using \LMVREGY~for Arabic words and \LMVRTGT~for English words) outperforms training the segmenter jointly (\LMVRJOINT). 
This setup gives the best performing model, 
referred to as \LMVRMT. 

Amongst the \textbf{supervised morphology-based segmenters}, the setup with \ATB~ 
is the best, which we refer to as \ATBMT. The finding is consistent with \citet{maiodeh}'s results.

For \textbf{unsupervised frequency-based segmenters}, using \BPE~to segment both source and target sides  
outperforms \LMVRMT~by +0.8 chrF2++ points and \ATBMT~by +1.6 chrF2++ points, which we refer to as \BPEMT. 
We observe that the ranking of these segmenters in MT performance is in reverse order compared to their ranking in segmentation task performance. We discuss this later in Section~\ref{sec:discussion}.



Most interestingly, contrary to \citep{maiodeh}, we find that applying {\BPE} on top of {\ATB}, 
which we refer to as  {\MIXMT}, slightly improves 
over {\BPEMT} but without statistical significance. However, {\MIXMT} outperforms {\ATBMT} and {\LMVRMT} with statistical significance.\footnote{We use Paired Significance Tests for Multi System Evaluation provided by SacreBLEU for the significance tests \url{https://github.com/mjpost/sacrebleu\#paired-bootstrap-resampling---paired-bs}.} 
We further investigate the effectiveness and statistical significance achieved by this approach in a learning curve with varying the training data size in Section~\ref{subsec:analysis}. 

Finally, we note that segmenting English words on the source and target sides consistently, while controlling all other conditions, is always advantageous, as shown in Table \ref{table:chrf2_scores_dev}.

\subsection{Analysis}
\label{subsec:analysis}
We further analyze the performance of the top \ac{mt} systems from each segmentation setup (\MIXMT, \BPEMT, \LMVRMT, and \ATBMT). 
We first look into the number of \ac{oov} tokens associated with each of the top-performing \ac{mt} systems to examine whether it has an impact on their final ranking. Secondly, we investigate whether the ranking of the systems is consistent across the different types of sentences. We evaluate the systems against 
varying morphological richness, English percentages, and CS types. Thirdly, 
we further investigate the effectiveness of applying BPE over ATB compared to using each segmenter on its own. We conduct this analysis across different CS types and sizes of training data.



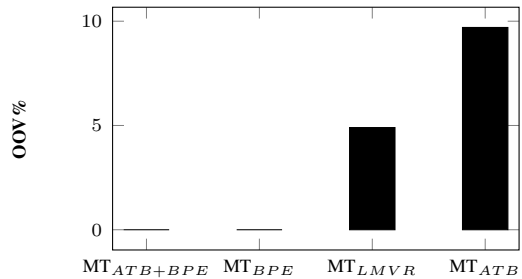
\begin{figure}[t]
  \begin{tikzpicture}
\begin{axis}[
    symbolic x coords={\MIXMT, \BPEMT,\LMVRMT,\ATBMT},
        ylabel = {\bf OOV\%},
        xlabel = {},
        font=\scriptsize,
         bar width=17pt,
          width=0.9\columnwidth,
        height=.3\textwidth,
        xticklabel style = {},
    xtick=data]
    \addplot[ybar,fill=black] coordinates {
        (\MIXMT,0)
        (\BPEMT,0)
        (\LMVRMT,4.90)
        (\ATBMT,9.70)
    };
\end{axis}
\end{tikzpicture}
    \caption{The percentage of the \ac{oov} words generated from each of the top-performing \ac{mt} systems from each segmentation setup on \textit{ArzEn}'s dev set.}
    \label{fig:oov_fig}
\end{figure}

\paragraph{OOV} To further study the reason behind {\BPEMT}~and {\MIXMT}~top performance, we observe if the 
 top-performing \ac{mt} 
 systems' ranking is linked with the percentage of \ac{oov} in the different \ac{mt} systems. As shown in Figure \ref{fig:oov_fig}, we find that for {\MIXMT}~and
{\BPEMT}~, the \ac{oov} percentage is 0\%. However, for 
{\LMVRMT}~and {\ATBMT}~, the percentage rises to 4.90\% and 9.70\%, respectively, which we believe contributes to worsening the \ac{mt} systems.

\paragraph{Evaluating Systems Under Different Sentence Categories} We evaluate the performance of the MT systems for sentences falling under different ranges of (i) morphological richness, (ii) percentage of CS English words, and (iii) sentence \ac{cs} types. Morphological richness of a sentence is calculated as the quotient of the number of tokens in the segmented sentence and unsegmented original sentence. As expected, the performance of all the \ac{mt} models decreases as the morphological richness increases and there is a boost in performance across all systems when the percentage of English words increases (see 
Appendix \ref{sec:appendix-eval-graphs}). We 
observe that the \MIXMT~and \BPEMT~perform the best across all ranges for the first two features. We then evaluate the performance of the MT systems across sentences according to CS types: purely monolingual Arabic, \ac{cs}, and \ac{cs} having \ac{mcs} \citep{hamed2022investigations,mager2019subword}. We observe that for all systems, the performance across CS sentences is higher than across monolingual Arabic sentences. 
We also observe that among \ac{cs} sentences, the performance is reduced in the case of morphologically code-switched sentences. 
We believe that the following two factors can be contributing to these results. Firstly, the complex 
\ac{mcs} constructions might impose challenges to the MT system. 
Secondly, we observe that the average length of \ac{mcs} sentences is higher than that of \ac{cs} sentences in general. This is partially due to the fact that the tokens in MCS words are space-separated during the data preprocessing step. We report that on average, \ac{cs} sentences contain 21.1 words (21.4 tokens), while \ac{mcs} sentences contain 25.0 words (26.3 tokens).
\paragraph{Further Investigating the Effectiveness of \MIXMT~ over \BPEMT~ and \ATBMT}
We study whether the ranking of \MIXMT, \BPEMT, and \ATBMT~is altered when going from a low-resource to an extreme low-resource setting across different sentence types. We achieve this 
by varying the \ac{mt} training data to 25\% and 50\% of its original size. The results are shown in Table~\ref{table:data_sizes}.

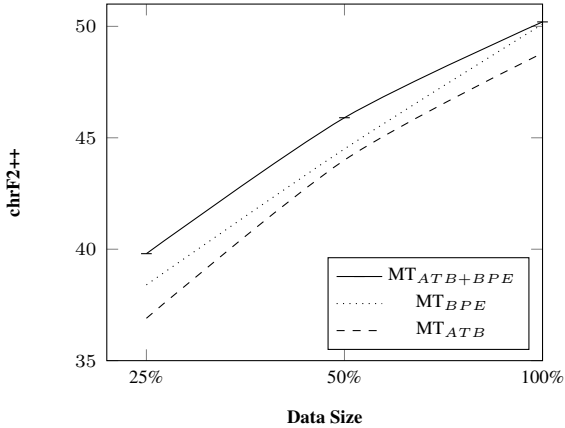
\begin{figure}[t]
 \begin{tikzpicture}
 
\begin{axis}[
width = 0.95\columnwidth,
    xlabel={\bf Data Size},
    ylabel={\bf chrF2++},
    xmin=8, xmax=30,
    ymin=35, ymax=51,
    xtick={10,20,30},
    xticklabels={25\%,50\%,100\%},  
    ytick={0,5,...,51},
    font=\scriptsize,
    legend pos=south east
         ]
\addplot[smooth,mark=-,black] plot coordinates {
    (10,39.8)
    (20,45.9)
    (30,50.2)
};
\addlegendentry{\MIXMT}

\addplot[smooth,color=black,dotted]
    plot coordinates {
        
        (10,38.4)
        (20,44.5)
        (30,50.1)
    };
\addlegendentry{\BPEMT}

\addplot[smooth,color=black, dashed]
    plot coordinates {
        
        (10,36.9)
        (20,44.0)
        (30,48.8)
    };
\addlegendentry{\ATBMT}
\end{axis}

    \end{tikzpicture}
    \caption{
    Demonstrates the effectiveness of applying \ac{bpe} on top of \ac{atb} segmentation (\MIXMT) as opposed to using either approaches separately (\BPEMT~ and \ATBMT), which is confirmed when reducing the amount of training data. Results are reported on \textit{ArzEn}'s dev set. 
    }
     \label{fig:analysis_line_chart}
     \vspace{-0.3cm}
\end{figure}

We observe that the effectiveness of the \MIXMT~ varies under constrained conditions. For monolingual Arabic sentences, when training the MT systems on 100\% of data, we see that \MIXMT~ is not statistically significant over \BPEMT~ and \ATBMT. Moreover, \MIXMT~ was outperformed by \BPEMT. However, when training with 25\% and 50\% of data, \MIXMT~ outperforms \BPEMT~ and \ATBMT~with statistical significance across all sentence types. We further exhibit this in Figure \ref{fig:analysis_line_chart} when all sentence categories are considered during analysis under different data sizes. This finding highlights the importance of combining morphology-based and frequency-based segmentations in extremely low-resource scenarios.

We also observe that across all data sizes, 
\ATBMT~performs the worst on \ac{cs} sentences. Our first hypothesis is that this is due to English words left unsegmented. However, results in Table~\ref{table:EMMA_scores} contradict this hypothesis. Our second hypothesis is that since \ATB~takes into consideration the context of the word prior to segmentation, the English words in the \ac{cs} sentences might break the flow of the sentence, hence negatively impacting the context of the word, thus worsening
the score.





\begin{table}[t]
\centering
\small
\setlength{\tabcolsep}{2pt}
\begin{tabular}{|l|l|c|c|c|c|}
\hline
  {\bf Size} &{\bf MT System} &  {\bf All}& {\bf EGY} & {\bf CS} & {\bf \ac{mcs}}\\\hline 
25\%  & {\bf \MIXMT} & \textbf{39.8 (1)} & \textbf{36.6 (1)}       & \textbf{40.6 (1)} & \textbf{40.0 (1)}   \\
      & {\bf \BPEMT}     & 38.4 (2) & 35.6 (3)       & 39.1 (2) & 38.5 (2)   \\
      & {\bf \ATBMT}     & 36.9 (3) & 35.9 (2)       & 37.0 (3) & 36.0 (3)   \\\hline \hline 
50\%  & {\bf \MIXMT} & \textbf{45.9 (1)} & \textbf{42.1 (1)}       & \textbf{46.8 (1)} & \textbf{46.4 (1)}   \\
      & {\bf \BPEMT}     & 44.5 (2) & 40.7 (3)       & 45.5 (2) & 44.8 (2)   \\
      & {\bf \ATBMT}     & 44.0 (3) & 41.4 (2)       & 44.7 (3) & 44.0 (3)   \\\hline \hline 
100\% & {\bf \MIXMT} & \textbf{50.2 (1)} & 44.4 (2)       & \textbf{51.5 (1)} & \textbf{51.3 (1)}   \\
      & {\bf \BPEMT}     & 50.1 (2) & \textbf{44.6 (1)}       & 51.3 (2) & 51.1 (2)   \\
      & {\bf \ATBMT}     & 48.8 (3) & 44.2 (3)       & 49.8 (3) & 49.4 (3)  \\\hline 
\end{tabular}
\caption{
We compare the results of the best performing \ac{mt} system (\MIXMT) which utilizes \ac{bpe} on top of \ac{atb} segmentation against the \ac{mt} systems that utilize \ac{bpe} ({\BPEMT}) or \ac{atb} ({\ATBMT}) only on \textit{ArzEn}'s dev set. We report chrF2++ results when training on 25\%, 50\%, and 100\% of the training data. Results are shown for different types of sentences: monolingual Egyptian Arabic (EGY), code-switched (CS), and morphologically code-switched (MCS), as well as all sentences (All). The ranking of the \ac{mt} systems with respect to each other is represented by the numbers between parentheses, where (1) is the best rank and (3) is the worst.
}\label{table:data_sizes}
\vspace{-0.3cm}
\end{table}

\paragraph{System Selection} As per our findings, \MIXMT~
is always the best choice 
across all sentence types in extreme low-resource settings. However, when training on 100\% of the data, \BPEMT~ improves slightly over \MIXMT~on monolingual Arabic sentences. Therefore, we create a system selection setup which 
uses both, \MIXMT~and \BPEMT, to investigate if it would lead to further improvements. 
In this setup, the \ac{cs} and monolingual English sentences are translated using {\MIXMT}, while monolingual Arabic sentences are translated using \BPEMT. Despite the hybrid system showing an overall improvement of +0.1 chrF2++ points over \MIXMT, the improvement is not statistically significant.

%

\pgfplotstableread[row sep=\\,col sep=&]{type&mix&bpe&lmvr&atb\\
all&50.2&50.1&48.9&48.8 \\
mono EGY&44.4&44.6&44.6&44.2 \\
cs&51.5&51.3&49.9&49.8\\
mcs&51.3&51.1&49.4&49.4\\
cs-mcs&51.8&51.6&50.5&50.5\\
}\langanalysis

\section{Discussion} \label{sec:discussion}
We revisit the RQs we outlined in our introduction.

\textbf{RQ1 - Which segmentation setup performs the best in the downstream MT task across different training sizes?} Results show that frequency-based segmentation applied on top of morphology-based segmentation outperforms the other segmentation techniques, with statistical significance on lower resource settings. The superiority of this approach is seen across sentences with varying morphological richness, percentage of English words, and across sentences with different CS types. 
%
%
We believe the strength of the combination is because it exploits complementarity of both methods. On one hand, supervised morphology-based segmenters bring in high correctness; however, they are not always robust, having high OOV rates. On the other hand, while BPE segmentation is not necessarily morphologically correct, it achieves high robustness. The robustness of BPE is consistent with the findings in \citet{banerjee2018meaningless}. 

\textbf{RQ2 - Does the effectiveness of the different segmenters in the MT task differ according to the \ac{cs} type of the source sentence?}
We observe that the effectiveness of the different segmenters on MT performance is consistent across two categories of CS sentences; those with and without MCS. However, when comparing their effectiveness on monolingual Arabic vs. \ac{cs} sentences, we observe that the rankings between segmenters are not consistent. In the case of constrained data size settings (25\% and 50\% of data), we observe a clear pattern where \ATBMT~outperforms \BPEMT~on monolingual sentences, while \BPEMT~ outperforms \ATBMT~on 
\ac{cs}. In the case of using 100\% of the training data, \MIXMT~outperforms \BPEMT~on \ac{cs} sentences; however, \BPEMT~outperforms \MIXMT~on monolingual Arabic sentences. Since 
our test and dev sets are dominated by \ac{cs} sentences (61.5\% and 63.8\%, respectively), we believe that the overall ranking is more greatly affected by the systems' performance on \ac{cs} sentences, thus reflecting the same ranking on the overall evaluation set as that across \ac{cs} sentences.

\textbf{RQ3 - Is there a correlation between a more morphologically correct segmentation and \ac{mt} performance?} For unsupervised morphology-based segmenters, a segmenter with a better segmentation EMMA F1 score also scores better in the downstream \ac{mt} task. However, we cannot hypothesize that a better segmentation score implies a better translation system, as counter examples exist. For example, while we notice that \ATB~gives the best segmentation in terms of EMMA F1 score, 
it does not outperform any of the top-performing \ac{mt} systems. We hypothesize that despite \ATB's capability of generating morphologically correct segmentations, it can generate infrequent morphemes due to the out-of-domain data which it is trained on. This may not only increase the sentence length which worsens
\ac{mt} performance as shown in \citet{mager2022bpe},
but may also be one of the contributing factors to the 9.70\% \ac{oov} percentage found in \ATBMT. On the contrary, \BPE~performs the worst in the segmentation task as we expect, since it is designated for agnostic-based segmentations; however, it surpasses the top-performing \ac{mt} models. We believe this is due to its capability to generate semi-correct segmentation and to reduce \ac{oov} rates.

\hide{
\begin{figure*}%
\centering

\begin{subfigure}{\textwidth}
\includegraphics[width=\textwidth]{images/f1.pdf}%
\caption{Sentence Length}%
\label{fig:sentence_length_analysis}%
\end{subfigure}\hfill%

\begin{subfigure}{\textwidth}
\includegraphics[width=\textwidth]{images/f2.pdf}%
\caption{Morphological Richness}%
\label{fig:morphological_richness_analysis}%
\end{subfigure}\hfill%

\begin{subfigure}{\textwidth}
\includegraphics[width=\textwidth]{images/f3.pdf}%
\caption{CS Percentage}%
\label{fig:ratios_analysis}%
\end{subfigure}%

\caption{Evaluating the performance of the best performing \ac{nmt} models against varying: sentence length, morphological richness, and CS percentage.}
\label{fig:MT_analysis}

\end{figure*}
}
\section{Conclusion and Future Work}\label{sec:conc}


In this paper, we study the impact of a comprehensive set 
of morphological and frequency-based segmentation methods on \ac{mt}, where the source is a code-switched text. The experiments are performed on code-switched Arabic-English to English. 
We found that the supervised morphological segmenter achieved the best performance on the segmentation task, followed by unsupervised morphological methods, and finally, unsupervised frequency-based. 
Afterward, we train 18 different \ac{mt} systems with different source and target side segmentations. We find that the rank of the segmenters is reversed, as
\ac{bpe}'s could not be outperformed (significantly) by any morphological-inspired segmentation method.
However, combining morphology-based and frequency-based segmentations has shown to give improvements, which are statistically significant in lower resource settings, where the training data size is reduced to 25\% and 50\%. 
For future work, we plan to apply our different \ac{mt} setups on other low resource and code-switched language pairs. Specifically, 
we plan to explore languages with different typologies, to study whether or not the relation between the data size and choice of the segmentation setup (frequency-based, morphology-based, or a mix) is based on morphological features and data size rather than the language itself. Moreover, we plan to extend our annotated dataset, \textit{\ac{arzenss}}, by adding further details to allow evaluating different schemes.

\section*{Limitations} \label{sec:limitation}

The first challenge we face in this work is the computational power needed to tune the Morfessor family segmenters. Therefore, in an attempt to overcome this challenge, for the Morfessor family, the choice of the optimal hyperparameters is dependent on the parent tool. For instance, the optimal hyperparameters for Morfessor are directly used in its FlatCat variant and the hyperparameters specific to FlatCat are then tuned. The same applies for LMVR which is a variant of FlatCat. Moreover, we cannot verify whether or not our results will hold for languages with different typologies, specifically those that are low resource and code-switched.
\hide{
 Finally, \ac{arzenss} accommodates only for one segmentation technique, \ac{atb}; hence other forms of correct morphological segmentations are perceived as incorrect and penalized by the EMMA score. For example, the word \<الكتب> \textit{Alktb} `the books' is segmented to \<كتب>\#\<ال> \textit{Al\#ktb} by some segmenters which is considered valid in segmentation schemes like D3. However, since we exploit \ac{atb} in \ac{arzenss} annotation, the EMMA system penalizes those segmenters. 
}
Therefore, the results of this research must be seen in light of these limitations.

\section*{Ethics Statement}
We could not find any potential harm that might derive from this work. However, we understand that translation as a whole can impact the cultural and social life of the people that use it. This has been used in the past in a harmful way, i.e., to spread colonial views \cite{mbuwayesango2018bible}. Therefore, we call the final user to use this work ethically. Regarding the annotation process, all manual annotations were made by a subset of the authors of this paper. Therefore, no hiring of external workers was necessary.


\section*{Acknowledgements}
We want to thank all the anonymous reviewers for their helpful comments
and suggestions. This project has benefited from financial support to Marwa Gaser and Manuel Mager by the DAAD (German Academic Exchange Service).

\bibliography{eacl2023}
\bibliographystyle{acl_natbib}

\newpage 
\clearpage

\appendix
\section{Data Preprocessing}
\label{sec:appendix-preprocessing}
We use the same preprocessing pipeline for all the corpora, where we start by removing any corpus-related annotations. Afterward, we remove URLs and emoticons, through \textit{tweet-preprocessor},\footnote{\url{https://pypi.org/project/tweet-preprocessor/}} remove trailing and leading spaces, and tokenize numbers. Finally, \textit{Moses Tokenizer}\footnote{\url{https://github.com/moses-smt/mosesdecoder/blob/master/scripts/tokenizer/tokenizer.perl}} is applied for tokenization and empty lines are removed from the parallel corpora. For LDC2017T07 \citep{ldc2017t07} and LDC2019T01 \citep{song2019}, some sentences have literal and intended translations for some words. Hence, we opt for one translation having all literal translations and another having all intended translations.
Once all the preprocessing steps are done, we concatenate the nine corpora collectively and pass the resulting training corpus to MADAMIRA \citep{madamira} to obtain two different supervised morphological segmentations of the corpus, namely \textit{{\fontfamily{qcr}\selectfont \ac{atb}\_BWFORM}} and \textit{{\fontfamily{qcr}\selectfont D3\_BWFORM}} which we discuss in Section \ref{subsec:segmentation_systems}. Additionally,  we obtain a raw training corpus by further tokenizing punctuation and removing emojis using MADAMIRA's D0 scheme \citep{zalmout2017optimizing}. Nonetheless, we normalize the Arabic letters \<ى> and \<أ> to  \<ي> and \<ا> respectively through CAMeL Tools \citep{camel} since  D0's output is not normalized.

\section{Segmenters' Hyperparameters}
\label{appx:segmenters}

\paragraph{Morfessor family}

Since all Morfessor family segmenters are morphology inspired, the hyperparameters are tuned on \textit{\ac{arzenss}}'s dev set. For \LMVRSRC~and \LMVRTGT~setting the vocabulary sizes to $64k$ and $16k$ respectively outperform $3k$, $5k$, $8k$, $16k$, $32k$, $100k$. For \LMVRJOINT~setting the vocabulary size to $32k$ outperforms $3k$, $5k$, $8k$, $16k$, $64k$, and $100k$. Meanwhile, For \LMVREGY~setting the vocabulary size to $64k$ outperforms $3k$, $5k$, $8k$, $16k$, $32k$, and $100k$. 


Table \ref{table:detailed_morf_family_hyper} shows the possible values used during the optimal hyperparameter search for each Morfessor tool. For Morfessor, FlatCat, and LMVR 18, 360, and 7 different segmentation models are generated. These are a result of the combination of the possible hyperparameter values. The hyperparameter combination which yields the highest EMMA score on \textit{\ac{arzenss}}'s dev set for each Morfessor tool is
used to segment the \ac{mt} training data. The best combination values are reported in Table  \ref{table:morfessor_hyperparams}.

\paragraph{MorphAGram}
Akin to the Morfessor family, we tune the hyperparameters on \textit{\ac{arzenss}}'s dev set and train two models: one on the source side and the other on the target side of the training parallel corpus which we refer to as \MORPHASRC~and \MORPHATGT, respectively (see Figure~\ref{fig:training_unsupervised_segmenters}). Tuning results show that setting the vocabulary size to $3k$ for \MORPHASRC~outperforms  $5k$, $8k$, $16k$, $32k$, and $50k$, while setting the vocabulary size to $50k$ for \MORPHATGT~outperforms  $5k$, $3k$, $8k$, $16k$, and $32k$. Nevertheless, it is worth noting that the vocabulary size on the target side is $<50k$ which shows that \MORPHATGT~performs the best when no segmentations are applied on the English words.

\paragraph{BPE}
Since \ac{bpe} is a segmentation technique that is designated for agnostic segmentation for \ac{mt} tasks, we tune the vocabulary size on \textit{ArzEn}'s dev set in an \ac{nmt} task. 
We apply a vocabulary size of $8k$, which outperforms $5k,16k,32k,64k$.

\begin{table}[t]
\small
\setlength{\tabcolsep}{2pt}
\centering
\begin{tabular}{|c|c|}
\hline
\multicolumn{2}{|c|}{\textbf{Segmenters Hyperparameters}}  \\
\cline{1-2}
\textbf{Hyperparameter} & \textbf{Values Bound}\\
\hline
\multicolumn{2}{|c|}{\it Morfessor}\\
\hline
-F & [0.003, 0.005, 0.007]\\
\hline
-d & [log, ones, none]\\
\hline
-a & [recursive, viterbi]\\
\hline
\multicolumn{2}{|c|}{\it FlatCat}\\
\hline
-p & [50, 60, 70, 80, 90, 100, 200, 300]\\
\hline
--min-perplexity-length & [1, 2, 3, 4, 5]\\
\hline
--min-shift-remainder & [1, 2, 3]\\
\hline
--length-threshold & [2, 3, 4]\\
\hline
\multicolumn{2}{|c|}{\it LMVR}\\
\hline
--lexicon-size & [3k, 5k, 8k, 16k, 32k, 64k, 100k]\\
\hline
\end{tabular}
\caption{The values bound we use during the best hyperparameter combination search for the Morfessor tools.} 
 \label{table:detailed_morf_family_hyper}
\end{table}

\begin{table*}[t]
\small
\centering
\begin{tabular}{|c|c|c|c|c|c|c|c|c|}
\cline{2-9}
\multicolumn{1}{c|}{} &\multicolumn{3}{c|}{\textbf{Morfessor}} &\multicolumn{4}{c}{\textbf{FlatCat}}&\multicolumn{1}{|c|}{\textbf{\ac{lmvr}}}\\
\hline
{\bf Data} & \textbf{-F} & \textbf{-d} & \textbf{-a} & \textbf{-p} &\textbf{--min} & \textbf{--min} &\textbf{--length} & \textbf{--lexicon}\\
&&&&&\textbf{-perplexity}&\textbf{-shift}&\textbf{-threshold}&\textbf{-size}\\
&&&&&\textbf{-length}&\textbf{-remainder}&&\\
\hline
src & 0.003&log&recursive &200&1&1&4&64k\\
\hline
tgt & 0.003&log&recursive&100&4&2&4&16k\\
\hline
src/egy & 0.007&log&recursive&300&1&1&2&64k\\
\hline
joint & 0.007&log&recursive&300&4&2&4&32k\\
\hline
\end{tabular}
\caption{The different hyperparameters used for each Morfessor family segmenter depending on whether the model is trained on the source (src), target (tgt), source without English, i.e., Egyptian, (src/egy), or source+target (joint) side(s).} 
 \label{table:morfessor_hyperparams}
\end{table*}

\section{Segmenters Performance Analysis}
\label{sec:appendix-SegmentersPerformanceAnalysis}
Table \ref{table:error_analysis_segmenters} shows the error analysis we perform on the segmenters with regards to over segmentation, under segmentation, or generating the correct number of segmentations per word.
\begin{table*}
\small
\centering
\begin{tabular}{lrrrrrrrrrr}
\cline{2-11}
\multicolumn{1}{c|}{} & \multicolumn{5}{c}{\bf EGY} & \multicolumn{5}{|c|}{\bf EN } \\
\hline
\multicolumn{1}{|c|} {\bf Segmenter}& \textbf{under} & \textbf{over} & \textbf{correct} &\textbf{seg.} &  { \textbf{unseg.}} &\multicolumn{1}{|c} {\textbf{under}} & \textbf{over} & \textbf{correct} &\textbf{seg}. & \multicolumn{1}{c|}{\textbf{unseg.}} \\
\hline
\multicolumn{1}{|c|}{raw} & 634 & 0    & 2,780 & 0   & 2,780  & \multicolumn{1}{|c}{71}  & 0    & 430 & 0  & \multicolumn{1}{c|}{430} \\
\hline
\hline
\multicolumn{1}{|c|}{\MORPHASRC} & 249	& 855	& 2,310 &	385	&1,925 & \multicolumn{1}{|c}{70}  & 45    & 386 &1  & \multicolumn{1}{c|}{385} \\

\multicolumn{1}{|c|}{\MORFSRC} & 466&	299 &	2,649&	148	&2,501 & \multicolumn{1}{|c}{15}  & 103    & 383 &42  & \multicolumn{1}{c|}{341} \\

\multicolumn{1}{|c|}{\FCSRC} & 592&	8&	2,814&	42 &	2,772 & \multicolumn{1}{|c}{56}  & 	7    & 438 &	15  & \multicolumn{1}{c|}{423} \\
			
\multicolumn{1}{|c|}{\LMVRSRC} & 520&	47&	2,847&	111	&2,736 & \multicolumn{1}{|c}{43}  & 	7    & 451 &	28  & \multicolumn{1}{c|}{423} \\

\hline
	\multicolumn{1}{|c|}{\MORPHATGT}&	634	&35&2,745&0&2,745&\multicolumn{1}{|c}{6}&148&347&65&\multicolumn{1}{c|}{282}\\
\multicolumn{1}{|c|}{\MORFTGT}	&0&3,150&264&3&261&\multicolumn{1}{|c}{21}&37 &443&49&\multicolumn{1}{c|}{394}\\

\multicolumn{1}{|c|}{\FCTGT} &	634&	0&	2,780&	0&	2,780&\multicolumn{1}{|c}{66}&	8&	427	&5&	\multicolumn{1}{c|}{422}\\
\multicolumn{1}{|c|}{\LMVRTGT}	&634	&0	&2,780	&0	&2,780 &\multicolumn{1}{|c}{23}&19	&459	&48&	\multicolumn{1}{c|}{411}\\
\hline

\multicolumn{1}{|c|}{\LMVRJOINT}&485&79&2,850&144&2,706&\multicolumn{1}{|c}{20}&32&449&51&\multicolumn{1}{c|}{398}\\
									
	\hline
	\hline
	\multicolumn{1}{|c|}{\BPE}&338&368&2,708&230&2,478&\multicolumn{1}{|c}{28}&132&341&30&\multicolumn{1}{c|}{311}\\
	\hline
	\hline 
	\multicolumn{1}{|c|}{\ATB}&38&62&3,314&581&2,733&\multicolumn{1}{|c}{71}&0&430&0&\multicolumn{1}{c|}{430}\\
	\multicolumn{1}{|c|}{\DT}&38&293&3,083&561&2,522&\multicolumn{1}{|c}{71}&0&430&0&\multicolumn{1}{c|}{430}\\
\hline

\end{tabular}
\caption{The table shows the number of under segmented words (under), over segmented words (over), and the number of cases where the segmenter generates the correct count of morphemes (correct) for English (EN) and Arabic (EGY) words in \textit{\ac{arzenss}} test set. Additionally, out of the correct count of morphemes (correct), we report the words which originally require segmentation (seg.) and those which do not (unseg.).}
 \label{table:error_analysis_segmenters}
\end{table*}

\section{MT Hyperparameters}
\label{sec:appendix-MThyperparameters}
The MT hyperparameters are shown in Table \ref{table:mtsystem_hyper}. We follow the FLORES hyperparameters for low-resource language pairs. The full train command can be found on FLORES GitHub.\footnote{\url{https://github.com/facebookresearch/flores/blob/6641ec0e23d173906dd2e01551a430884b1dba31/floresv1/README.md\#train-a-baseline-transformer-model}} The training time for \ac{mt} model the training time is exhibited in Table \ref{table:training_time_mt}.

\begin{table}[ht]
\small
\centering
\begin{tabular}{|c|c|}
\hline

\textbf{Hyperparameter} & \textbf{Value} \\
 \hline
 encoder-layers & 5\\
 decoder-layer & 5\\
 encoder-embed-dim & 512\\
decoder-embed-dim & 512 \\
encoder-ffn-embed-dim & 2 \\
decoder-ffn-embed-dim & 2\\
dropout & 0.4\\
attention-dropout &0.2\\
relu-dropout & 0.2 \\
weight-decay&0.0001\\
label-smoothing& 0.2\\
warmup-updates & 4000 \\
warmup-init-lr & 1e-9\\

 \hline
\end{tabular}
\caption{FLORES hyperparameters for low-resource language pairs.}
 \label{table:mtsystem_hyper}
\end{table}

\begin{table}[t]
\centering
\small
\setlength{\tabcolsep}{1pt}
\begin{tabular}{|c|c|c|r|}
\hline
\multicolumn{3}{|c|}{{\bf Segmentation}} & \multicolumn{1}{c|}{\bf Training} \\
 \cline{1-3}
\multicolumn{2}{|c|}{{\bf Source}} & {\bf Target} & \multicolumn{1}{c|}{\bf Time} \\  \cline{1-3}
 {\bf EGY} & {\bf EN} & {\bf EN}  & \multicolumn{1}{c|}{\textit{\textbf{(seconds)} }}    \\ \hline\hline
\multicolumn{2}{|c|} \RAW & \RAW & 13,522   \\
\hline\hline
\multicolumn{4}{|c|}{\it Unsupervised Morphology-based Segmenters}\\\hline
\multicolumn{2}{|c|}\MORPHASRC &  \MORPHATGT & 24,731 \\
\multicolumn{2}{|c|}\MORFSRC  & \MORFTGT  & 18,916\\
\multicolumn{2}{|c|}\FCSRC  &   \FCTGT  & 18,225\\
\multicolumn{2}{|c|}\LMVRSRC  &   \LMVRTGT & 4,476\\
\multicolumn{2}{|c|}\LMVRJOINT  &   \LMVRJOINT & 18,019\\\hline
\multicolumn{1}{|c|}\LMVREGY  &  \LMVRTGT &  \LMVRTGT & 22,462\\
\multicolumn{1}{|c|}\LMVRSRC  &  \LMVRTGT &  \LMVRTGT &4,181\\
\multicolumn{1}{|c|}\LMVREGY  &  \LMVRSRC &  \LMVRTGT & 4,526\\
\hline\hline
\multicolumn{4}{|c|}{\it Frequency-based Segmenters}\\\hline

\multicolumn{2}{|c|}\BPE &  \BPE & 18,279 \\
\multicolumn{2}{|c|}\BPE &  \RAW & 23,193\\
\multicolumn{2}{|c|}\RAW &  \BPE & 17,905\\
 \hline\hline
 \multicolumn{4}{|c|}{\it Supervised Morphology-based Segmenters}\\\hline

{\ATB}  &  {\RAW} &  {\RAW} & 18,280\\
{\DT}  &  \RAW &  \RAW & 18,519\\

\hline\hline
\multicolumn{4}{|c|}{\it Combination Segmenters}\\\hline
{\ATB+\BPE}  &  \BPE &  \RAW & 17,629\\
{\ATB+\BPE}  &  \BPE &  \BPE &27,088 \\
{\DT+\BPE}  &  \BPE &  \RAW &  24,256\\
{\DT+\BPE}  & \BPE &  \BPE & 23,611\\
\hline
\end{tabular}
\caption{The training time in seconds of our different \ac{nmt} systems.}
\label{table:training_time_mt}
\end{table}

\section{Evaluating Systems Under Different Sentence Categories}
\label{sec:appendix-eval-graphs}
\begin{figure*}[t]
\centering
\begin{subfigure}{\textwidth}
    \includegraphics[width=\textwidth]{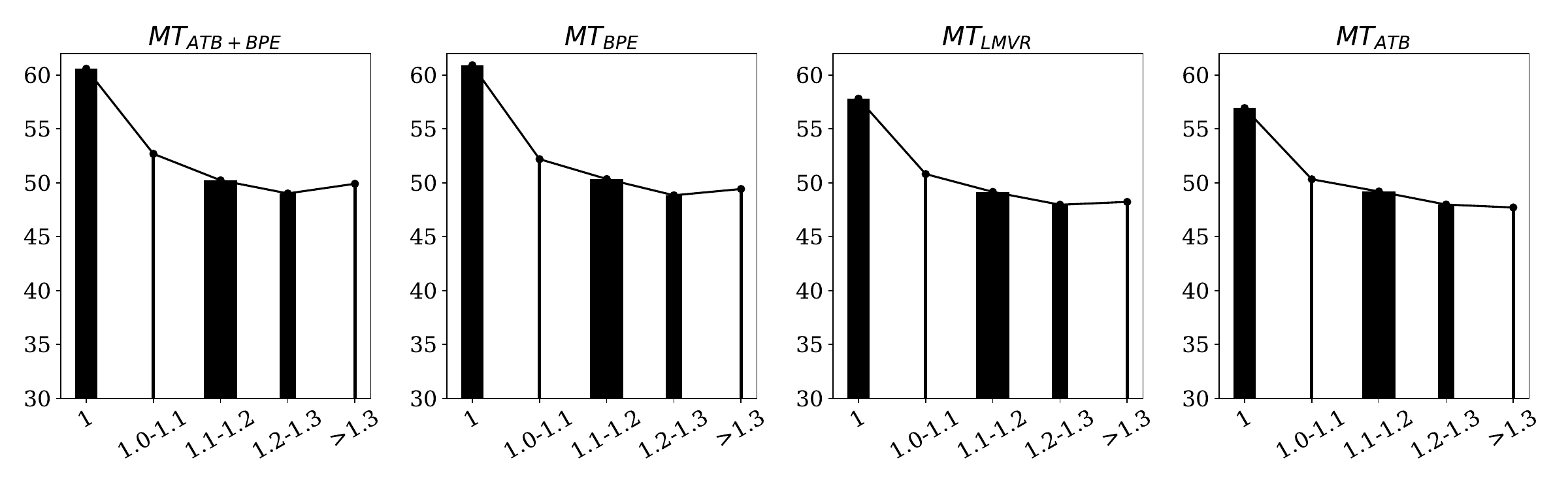}
    \caption{Morphological Richness}
    \label{fig:morphological_richness_analysis}
\end{subfigure}
\hfill
\begin{subfigure}{\textwidth}
    \includegraphics[width=\textwidth]{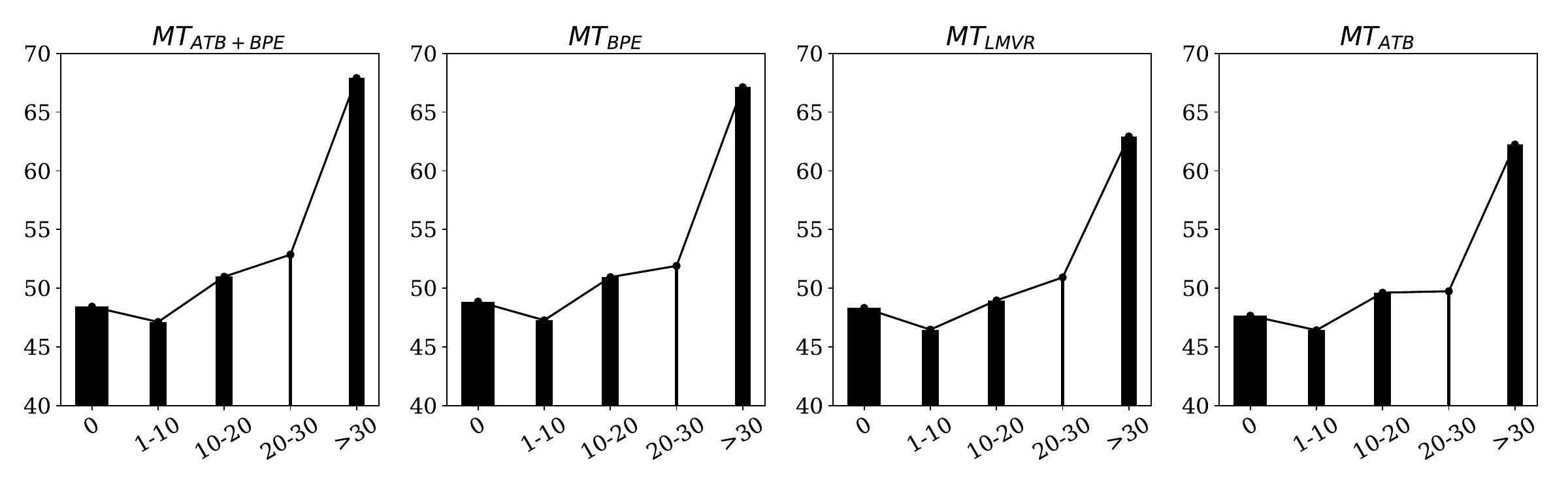}
    \caption{\% English Words}
    \label{fig:english_percentage_analysis}
\end{subfigure}
\caption{
The average chrF2++ scores for the top performing \ac{mt} systems from each segmentation setup across sentences with various (a) morphological richness ratios and (b) percentage of English words in \textit{ArzEn}'s dev set. Morphological richness of a sentence is calculated as the quotient of the number of tokens in the segmented sentence and unsegmented original sentence. The bar width is indicative of the number of sentences in each bin.}
\label{fig:morphologicalRichness_englishPercentage_analyses}
\end{figure*}

Figure \ref{fig:morphologicalRichness_englishPercentage_analyses} shows the performance of the top \ac{mt} systems from each segmentation setup across sentences of different morphological richness ratios and different percentages of English words in \textit{ArzEn}'s dev set. Results show that there is a general decrease in performance as the morphological richness increases. However, as the percentage of English words in the sentences increases, the performance of the systems generally improves. It is also shown that {\MIXMT} and {\BPEMT} achieve overall comparable performances and outperform the other systems.

\end{document}